\documentclass[10pt,twocolumn,letterpaper]{article}

\usepackage[pagenumbers]{cvpr}

\usepackage{cuted} 

\usepackage{currfile} 

\usepackage{caption} 

\usepackage[T1]{fontenc}

\usepackage{algorithm}
\usepackage{algpseudocode}

\usepackage{placeins}
\definecolor{cvprblue}{rgb}{0.21,0.49,0.74}
\usepackage[pagebackref,breaklinks,colorlinks,allcolors=cvprblue]{hyperref}

\def\confName{CVPR}
\def\confYear{2027}

\title{Wearable Logo, Silent Hijack: Stealthy Cross-Modal Semantic Attacks on Visible--Infrared Vision-Language Models}

\author{Qike Zhang, Jiaju Han, Xuelian Shi, Xiang Chen, Dehe Zhao, Banglong Nong, Wenhao Zheng,\\
Lei Wang, Jiahuan Long, Jiujiang Guo, Chengyin Hu}
\hypersetup{pdfauthor={Qike Zhang, Jiaju Han, Xuelian Shi, Xiang Chen, Dehe Zhao, Banglong Nong, Wenhao Zheng, Lei Wang, Jiahuan Long, Jiujiang Guo, Chengyin Hu}}

\begin{document}
\maketitle
\begin{abstract}
Logos appear naturally on clothing and daily necessities, where their recognizable shapes and semantic associations make them familiar visual cues.
This familiarity helps them blend into real-world environments and remain visually inconspicuous.
We ask whether an existing logo can serve as a carrier for coordinated targeted semantic attacks on visible--infrared (VIS--IR) vision-language models (VLMs) without changing its original outline.
This setting requires expressive adversarial variation within a fixed shape while accounting for different appearance characteristics and model responses across modalities.
We propose \textnormal{Logo-guided Semantic Attack (LoSA)}, a score-based black-box framework that optimizes a logo's interior appearance and placement while preserving its silhouette.
LoSA uses a contour field and low-frequency graph spectral modes to structure the interior color layout, a frozen appearance adapter to couple VIS and IR appearances, and particle swarm optimization to jointly optimize shared design parameters.
Experiments across four datasets demonstrate that LoSA consistently outperforms the evaluated baselines in simultaneous Top-1 targeted success.
Cross-architecture experiments, visual naturalness ratings, and captioning/VQA transfer analyses further characterize its effectiveness and limitations.
These findings reveal that familiar visual symbols from real-world environments can serve as shape-preserving carriers for coordinated targeted semantic attacks across imaging modalities.
\end{abstract}

\begin{figure}[t]
    \centering
    \includegraphics[width=\linewidth]{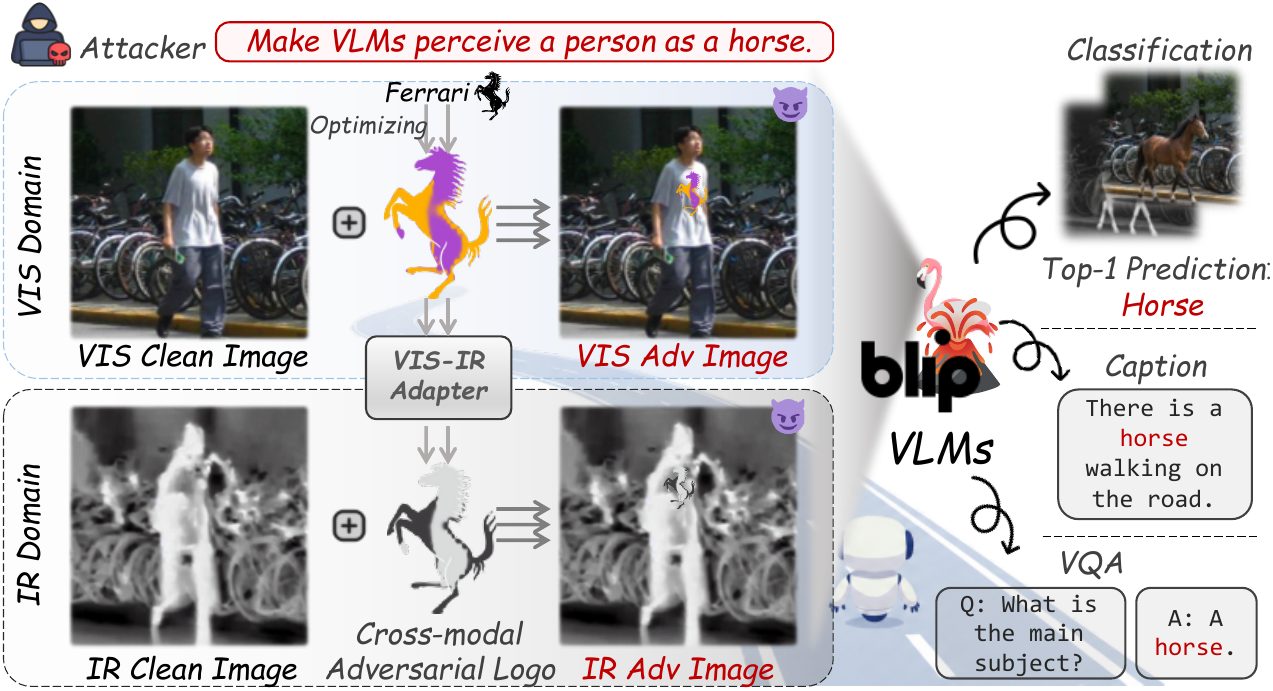}
    \caption{\textbf{Conceptual overview of LoSA.} A familiar logo's interior fills are optimized while its outline is retained. An adapter couples its VIS and IR appearances, and the logos are overlaid to induce the same target prediction.}
    \label{fig:losa_overview}
\end{figure}

\section{Introduction}
\label{sec:intro}

Logos are a common part of clothing, accessories, and daily necessities.
Familiar logos have recognizable outlines and established visual identities.
For humans, logos carry real-world meanings beyond their visual outlines, such as associations with particular brands. For vision--language models (VLMs)~\cite{radford2021clip,li2023blip2,liu2023improved}, their contextual plausibility and shape-based visual content may jointly make them a source of competing semantic cues.
This raises a question: can a familiar logo serve as an adversarial carrier that steers both visible and infrared VLMs toward the same prescribed target while remaining recognizable to humans as an ordinary logo?

Existing adversarial attack studies have investigated visual plausibility and attacks in visible--infrared dual-modal settings. Naturalistic Patch and AdvLogo use generative priors to produce meaningful patterns~\cite{hu2021naturalistic,miao2025advlogo}.
For VLMs, SLANT mines target-associated logos, while Web Artifact Attacks searches for graphical cues that influence recognition~\cite{qraitem2024slant,qraitem2025webartifacts}.
Visible--infrared (VIS--IR) patch methods also coordinate attacks on detectors across spectra~\cite{wei2023unipatch,hu2023touap,long2025cdupatch}. 
However, these approaches still face limitations in visual naturalness or in allowing attackers to freely specify the target prediction.
We therefore investigate targeted attacks on VIS--IR VLMs that preserve a supplied logo's outline and optimize its interior appearance and placement. Our objective is to steer both branches to the same attacker-specified category, beyond merely causing incorrect predictions.

Two coupled challenges arise in this setting.
First, preserving the supplied outline restricts the domain available for appearance optimization.
The interior must remain sufficiently expressive to change model predictions within that fixed shape.
Second, one design must operate across VIS and IR despite different appearance characteristics and model responses.
These requirements are linked. The same limited appearance space must supply effective cues for both branches, and success requires both to predict the prescribed target.

To address these challenges, we introduce Logo-guided Semantic Attack (LoSA) around one design principle: preserve the supplied logo outline while optimizing its interior appearance for coordinated targeting. \Cref{fig:losa_overview} illustrates the concept of our attack.
A boundary-derived contour field helps preserve the logo's semantic identity during interior appearance optimization, while low-frequency graph spectral modes enable structured variations within its fixed outline.
A compact renderer produces the VIS texture, and a frozen appearance adapter derives its IR counterpart.
Particle swarm optimization jointly optimizes shared appearance and placement parameters under a paired targeted objective.

Our contributions are threefold:
\begin{itemize}
    \setlength{\itemsep}{2pt}
    \item We investigate shape-preserving, logo-guided targeted semantic attacks on VIS--IR VLMs, examining whether an existing visual symbol can steer both branches toward the same attacker-specified category.
    \item We propose LoSA, which couples a contour-guided, graph-spectral representation of interior appearance with a frozen VIS-to-IR mapping and joint black-box optimization in a shared design space.
    \item Digital experiments on four datasets demonstrate improved simultaneous targeted success over the evaluated patch baselines. Further analyses characterize visual plausibility on selected scenes, transfer beyond classification, and sensitivity to architectures and removal defenses.
\end{itemize}

\section{Related Work}
\label{sec:related_work}

\noindent\textbf{Adversarial attacks on vision--language models.}
Image--text alignment enables zero-shot recognition and supports generative tasks such as captioning and visual question answering~\cite{radford2021clip,li2023blip2}.
Targeted visual attacks can alter predictions and downstream textual responses~\cite{zhao2023attackvlm,zhang2025anyattack}.
Chain of Attack uses iterative multimodal semantic updates to improve adversarial transferability~\cite{xie2025chainattack}, while V-Attack targets disentangled value features to steer selected concepts toward prescribed semantic targets~\cite{nie2026vattack}.
LoSA focuses on a fixed logo carrier and requires the same attacker-specified category in both VIS and IR branches.

\smallskip
\noindent\textbf{Naturalistic and logo-based adversarial attacks.}
Generative patch methods use image priors to balance attack effectiveness and visual plausibility.
Naturalistic Patch searches the image manifold of a pretrained generative adversarial network (GAN)~\cite{hu2021naturalistic}, while AdvLogo optimizes diffusion representations within a chosen semantic space~\cite{miao2025advlogo}.
Logo-shaped carriers also have a history in detector attacks: 3D Adversarial Logos maps adversarial textures onto human meshes representing clothed people~\cite{wang2020adversariallogos}.
For VLMs, SLANT searches logo banks for spurious target associations, while Web Artifact Attacks retrieves graphical artifacts and optimizes their placement~\cite{qraitem2024slant,qraitem2025webartifacts}.
In LoSA, the supplied outline instead defines a fixed optimization domain for interior color layouts shared across VIS and IR.

\smallskip
\noindent\textbf{Visible--infrared adversarial attacks.}
VIS--IR patch attacks account for the different imaging characteristics of the two spectra.
Unified Patch optimizes patch shape~\cite{wei2023unipatch}, TOUAP combines infrared shape optimization with visible texture search~\cite{hu2023touap}, and CDUPatch couples visible and infrared textures through an RGB-to-infrared adapter~\cite{long2025cdupatch}.
Thermally activated adversarial clothing further integrates dual-spectrum evasion into garments~\cite{long2026thermalclothing}.
Infrared vision--language modeling extends the task scope to image--text semantic understanding~\cite{cao2025irgpt}.
LoSA requires coordinated semantic redirection to the same chosen category in both VLM branches, extending the objective beyond detector evasion.

\begingroup
\setlength{\abovedisplayskip}{5pt plus 1pt minus 1pt}
\setlength{\belowdisplayskip}{5pt plus 1pt minus 1pt}
\setlength{\belowdisplayshortskip}{4pt plus 1pt minus 1pt}
\setlength{\textfloatsep}{10pt plus 2pt minus 2pt}
\setlength{\dbltextfloatsep}{10pt plus 2pt minus 2pt}
\setlength{\floatsep}{6pt plus 2pt minus 2pt}
\setlength{\dblfloatsep}{8pt plus 2pt minus 2pt}
\setlength{\intextsep}{8pt plus 2pt minus 2pt}
\captionsetup{skip=4pt}
\section{Method}
\label{sec:method}

\begin{figure*}[t]
    \centering
    \includegraphics[width=\textwidth]{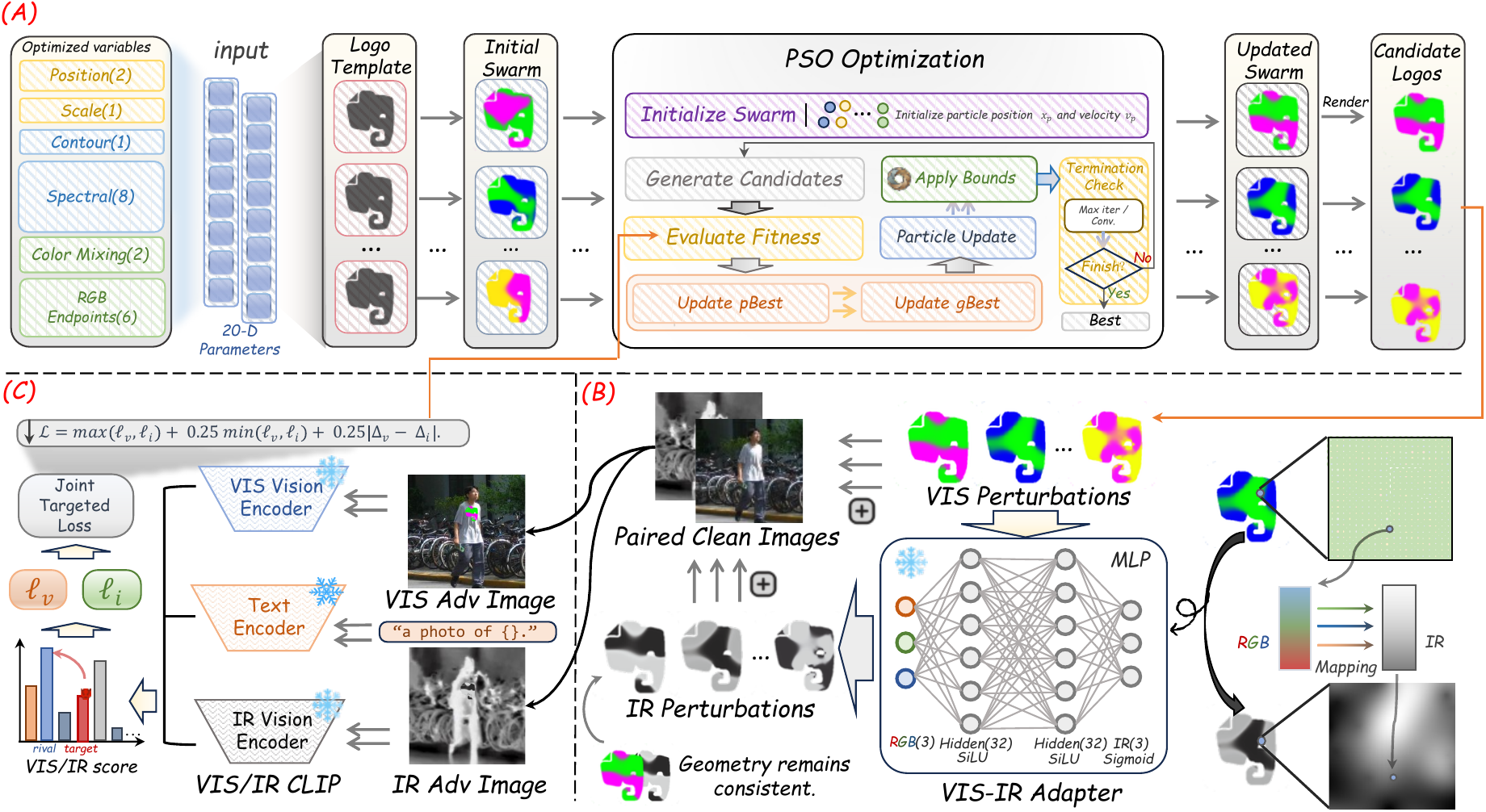}
    \caption{\textbf{LoSA framework.} (A) PSO optimizes parameters in the shared 20-dimensional design space and updates candidate logos using their fitness. (B) A frozen appearance adapter maps rendered RGB textures to IR appearances before digital compositing. (C) Frozen VIS and IR encoders provide class scores for the joint targeted objective; no model gradients are used to update the logo parameters.}
    \label{fig:losa_framework}
\end{figure*}

We present Logo-guided Semantic Attack (LoSA), which seeks the same target label on paired visible (VIS) and infrared (IR) images, as illustrated in \cref{fig:losa_framework}.
The selected logo's outline and associated concept anchor the color optimization, and a field defined on its foreground controls the visible color layout.
We render that layout on each patch grid and use a frozen adapter to derive the IR appearance.
A joint objective guides optimization of the shared design parameters.

\subsection{Problem formulation and threat model}
\label{sec:threat_model}

Let $(x_v,x_i)$ be a VIS--IR image pair with supplied object boxes $(b_v,b_i)$.
The attack also takes a logo asset and a target category $t$ from a fixed candidate set $\mathcal C$.
We assume score-based black-box access to each frozen vision--language model.
Each evaluation returns the complete class-score vector for that branch.

For branch $m\in\{v,i\}$, let $f_m(x)$ and $g_m(p_k)$ be the unit-normalized image and text embeddings from its encoders.
We insert the category name into the prompt template \mbox{\texttt{a photo of a \{\}}} to form $p_k$.
Following CLIP~\cite{radford2021clip}, the class score is their cosine similarity,
\begin{equation}
    s_{m,k}(x)=f_m(x)^\top g_m(p_k).
    \label{eq:class_score}
\end{equation}

The shared design $\theta$ includes three normalized geometry coordinates $\xi$ for horizontal position, vertical position, and size.
We preserve the logo's proportions and use the object box to bound its center optimization.
The patch may extend beyond the box, while its full rectangle remains inside the image canvas.
Let $\bar P_m$ and $\bar M_m$ denote the branch texture and its resized alpha mask after placement on the canvas.
We composite the adversarial input as
\begin{equation}
    x_m^{\mathrm{adv}}(\theta)=(1-\bar M_m)\odot x_m
    +\bar M_m\odot\bar P_m.
    \label{eq:alpha_compositing}
\end{equation}
The canvas mask is zero outside the patch rectangle.

Let $\hat y_m(\theta)$ denote the top-ranked category on $x_m^{\mathrm{adv}}(\theta)$.
An attack succeeds when both branches predict the target,
\begin{equation}
    \hat y_v(\theta)=\hat y_i(\theta)=t.
    \label{eq:target_success}
\end{equation}
Only image pairs with consistent clean Top-1 predictions across the VIS and IR branches are eligible for attack optimization.

\subsection{Contour-preserving visible logo rendering}
\label{sec:spectral_logo}

\begin{figure*}[t]
    \centering
    \includegraphics[width=\textwidth]{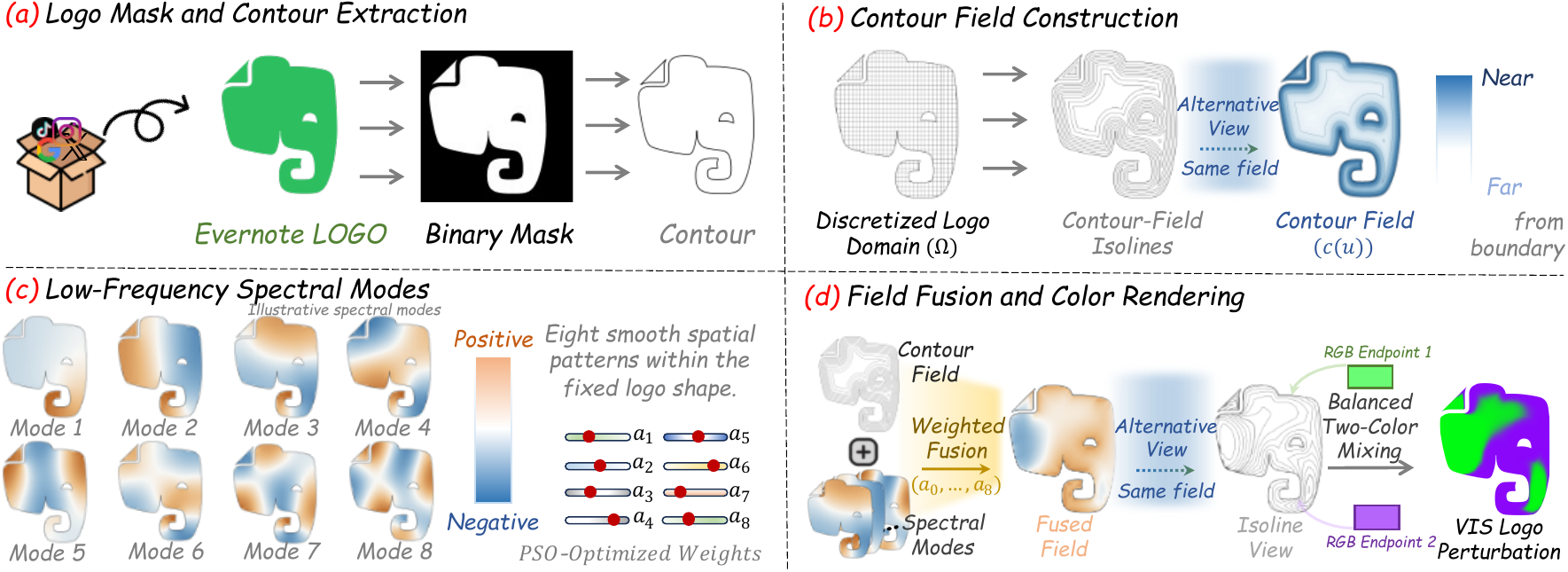}
    \caption{\textbf{Contour-preserving logo rendering.} (a) The supplied logo defines a fixed foreground domain. (b) A boundary-derived contour field anchors the interior layout to that shape. (c) Eight graph spectral modes provide structured variation within the domain; their illustrations are schematic. (d) Optimized field weights and two RGB endpoints produce the visible texture through balanced color mixing. Isoline views depict the same scalar fields, rather than additional rendering stages.}
    \label{fig:logo_rendering}
\end{figure*}

\Cref{fig:logo_rendering} illustrates how the fixed logo shape supports the contour field, spectral variations, and two-color filling described below.

A scalar field controls how two RGB endpoint colors are distributed within the fixed logo shape.
The contour field anchors the fill to the logo's original shape and associated semantic cues, while eight spectral modes, two RGB endpoints, and two mixing controls provide flexibility in spatial layout and color allocation.

\paragraph{Fixed logo outline.}
We rasterize the logo's alpha channel and crop the transparent margins.
The resulting mask preserves the outer contour and any transparent regions inside it.
We resize and threshold the alpha mask on a $32\times32$ grid to define the foreground domain $\Omega$.
This domain supports both the contour field and the spectral modes.

\paragraph{Contour field.}
Let $d(u,\partial\Omega)$ be the Euclidean distance in grid pixels from a foreground pixel $u$ to the boundary.
The contour field is
\begin{equation}
    c(u)=\exp\!\left[-d(u,\partial\Omega)/2\right],
    \qquad u\in\Omega.
    \label{eq:contour_field}
\end{equation}
Its values decay from the boundary toward the interior.
Because the boundary is inherited from the original logo, the field ties color allocation to the shape carrying its associated concept.

\paragraph{Spectral modes and field composition.}
We connect four-neighbor foreground pixels and form the graph Laplacian $L=D-A$, where $D$ and $A$ are the degree and adjacency matrices.
The eight eigenvectors with the smallest positive eigenvalues define modes $\phi_1,\ldots,\phi_8$~\cite{shuman2013graphs}.
Selecting low eigenvalues favors variations with small changes between adjacent foreground pixels.
We form the combined field with the coefficient vector $a$,
\begin{equation}
    h_{a}(u)=\tanh\!\left(
    \frac{a_0c(u)+\sum_{j=1}^{8}a_j\phi_j(u)}{3}
    \right).
    \label{eq:spectral_field}
\end{equation}
The coefficient $a_0$ weights the boundary contribution, while the remaining eight vary the interior layout within the same semantic shape.
We extend the field by zero outside $\Omega$.

\paragraph{Color filling.}
At the selected patch size, we resize the alpha mask to $M$ and bicubically interpolate the field.
We clip the interpolated values to $[-1,1]$ to obtain $z$.

The color transition is centered on a quantile of $z$ over pixels $V=\{u:M(u)>0.1\}$.
The bias $\beta$ sets the quantile level $\rho=0.4+0.2\sigma(\beta)$, where $\sigma$ is the sigmoid.
We denote the resulting quantile by $q$ and use the standard deviation of $z$ over $V$, floored at $10^{-3}$, as the scale $\nu$.
The gain $\gamma$ controls the sharpness of the transition through the logit $\eta$ and mixing weight $w$,
\begin{equation}
    \begin{aligned}
        \eta(u) &= (4+2\gamma)\,[z(u)-q]/\nu,\\
        w(u) &= \sigma\!\left(\operatorname{clip}(\eta(u),-20,20)\right).
    \end{aligned}
    \label{eq:balanced_mixing}
\end{equation}
We use these weights to interpolate between the two optimized RGB endpoints $c_0,c_1\in[0,1]^3$,
\begin{equation}
    R(u)=(1-w(u))\,c_0+w(u)\,c_1.
    \label{eq:rgb_texture}
\end{equation}
The texture $R$ and mask $M$ specify the visible patch before compositing.

\subsection{Visible-to-infrared logo mapping}
\label{sec:cross_modal_rendering}

We apply the RGB renderer to the VIS and IR patch grids with the same appearance parameters, obtaining textures $R_v$ and $R_i$.
A frozen adapter $A_\psi$ converts each mixed RGB pixel on the IR grid to a three-channel IR appearance surrogate.
The two patches are
\begin{equation}
    P_v(u)=R_v(u),\qquad P_i(u)=A_\psi\!\left(R_i(u)\right).
    \label{eq:adapter_rendering}
\end{equation}

The adapter has two hidden layers of width 32 with SiLU activations and a three-channel sigmoid output.
We fit it using RGB swatches and their DiffV2IR~\cite{ran2025diffv2ir} translations.
For each swatch pair, the channelwise medians of the interior pixels provide the source and target triplets.
We minimize the mean squared error between the predicted and target triplets before running the attack.

\subsection{Joint cross-modal targeted objective}
\label{sec:joint_objective}

We score the composited pair from \cref{eq:alpha_compositing} and measure the target's score relative to its strongest rival in each branch.
For a candidate design, abbreviate its scores as $s_{m,k}$ and let $s_{m,\mathrm r}$ be the largest score among classes other than $t$.
Let $\sigma_m$ denote the population standard deviation over all class scores in that branch.
The standardized target margin is
\begin{equation}
    \Delta_m=\frac{s_{m,t}-s_{m,\mathrm r}}{\sigma_m+\epsilon},
    \qquad \epsilon=10^{-6}.
    \label{eq:standardized_margin}
\end{equation}
The common score mean cancels when we subtract the standardized scores, leaving the gap scaled by the branch's score spread.

The branch loss \mbox{$\ell_m=\log(1+e^{-\Delta_m})$} decreases as the target margin grows.
Let $\ell_{\max}$ and $\ell_{\min}$ be the larger and smaller of the two branch losses.
The joint objective is
\begin{equation}
    \mathcal L(\theta)=\ell_{\max}+\lambda_e\ell_{\min}
    +\lambda_b|\Delta_v-\Delta_i|,
    \label{eq:joint_loss}
\end{equation}
with $\lambda_e=\lambda_b=0.25$.
The larger loss prioritizes the branch with the smaller target margin, and the smaller loss rewards progress in the other branch.
The final term penalizes a gap between the two margins.

\subsection{Particle swarm optimization}
\label{sec:pso}

Each particle stores the shared design
\begin{equation}
    \theta=(\xi,a,\gamma,\beta,c_0,c_1)
    \in\mathbb R^{20}.
    \label{eq:parameter_vector}
\end{equation}
Its 20 coordinates cover three geometry variables, nine field coefficients, two mixing controls, and six color values.
We minimize \cref{eq:joint_loss} for each image pair with particle swarm optimization (PSO)~\cite{kennedy1995pso}.
The RGB endpoints start from eight preset pairs and their reversals, with Gaussian perturbations.
Both endpoints remain continuously adjustable throughout optimization.

Let $T_{\max}$ denote the maximum number of PSO iterations; we set $T_{\max}=200$ in our experiments.
Each iteration evaluates both branches for all 20 particles.
Velocity updates combine inertia with attraction to the particle's personal best and the swarm best, subject to velocity and parameter bounds.
A candidate qualifies for early stopping if it satisfies \cref{eq:target_success} with both standardized margins at least $0.05$.
Algorithm~\ref{alg:losa} takes the lowest-loss candidate from the first iteration that meets the stopping rule.
If the $T_{\max}$-iteration budget expires, it selects the best candidate found by the swarm.
We score the selected pair once more and assess final success using \cref{eq:target_success}.

\begin{algorithm}[H]
    \caption{LoSA optimization for one image pair}
    \label{alg:losa}
    \begin{algorithmic}[1]
        \Require Eligible VIS--IR pair, boxes, logo mask, target $t$
        \State Construct the contour field and spectral modes
        \State Initialize the bounded swarm
        \For{$r=1,\ldots,T_{\max}$}
            \State Render the adversarial pair for each particle
            \State Evaluate \cref{eq:joint_loss} and update personal bests
            \State $\theta^\star\gets$ swarm best
            \State $S\gets$ particles satisfying the early-stopping rule
            \If{$S\ne\varnothing$}
                \State $\theta^\star\gets\arg\min_{\theta\in S}\mathcal L(\theta)$
                \State \textbf{break}
            \EndIf
            \State Update and clip velocities and positions
        \EndFor
        \State Render and score the selected pair
        \State \Return $\theta^\star$ and the resulting image pair
    \end{algorithmic}
\end{algorithm}

\FloatBarrier
\endgroup
\begingroup
\setlength{\textfloatsep}{4pt plus 1pt minus 1pt}
\setlength{\dbltextfloatsep}{4pt plus 1pt minus 1pt}
\setlength{\floatsep}{4pt plus 1pt minus 1pt}
\setlength{\dblfloatsep}{5pt plus 1pt minus 1pt}
\setlength{\intextsep}{5pt plus 1pt minus 1pt}
\captionsetup{skip=4pt}
\section{Experiments}
\label{sec:experiments}

\begin{table*}[t]
    \centering
    \caption{\textbf{Zero-shot attack effectiveness across datasets.}
    Rates (\%) on OpenAI CLIP ViT-L/14. Datasets: RGBDT500,
    LasHeR (controlled subset), COCO, and CrowdHuman.
    Both requires the outcome in both branches on the same pair.
    @1/@3 denote target Top-1/Top-3 matches. Best values are bold.}
    \label{tab:classification_effectiveness}
    \small
    \setlength{\tabcolsep}{2.2pt}
    \resizebox{\textwidth}{!}{%
    \begin{tabular}{@{}lccccc*{3}{@{\hspace{7pt}}ccccc}@{}}
        \toprule
        & \multicolumn{5}{c}{RGBDT500}
        & \multicolumn{5}{c}{LasHeR}
        & \multicolumn{5}{c}{COCO}
        & \multicolumn{5}{c}{CrowdHuman} \\
        \cmidrule(lr){2-6} \cmidrule(lr){7-11}
        \cmidrule(lr){12-16} \cmidrule(lr){17-21}
        & \multicolumn{2}{c}{TASR@1} & \multicolumn{3}{c}{Both}
        & \multicolumn{2}{c}{TASR@1} & \multicolumn{3}{c}{Both}
        & \multicolumn{2}{c}{TASR@1} & \multicolumn{3}{c}{Both}
        & \multicolumn{2}{c}{TASR@1} & \multicolumn{3}{c}{Both} \\
        \cmidrule(lr){2-3} \cmidrule(lr){4-6}
        \cmidrule(lr){7-8} \cmidrule(lr){9-11}
        \cmidrule(lr){12-13} \cmidrule(lr){14-16}
        \cmidrule(lr){17-18} \cmidrule(lr){19-21}
        Method
        & VIS & IR & @1 & @3 & ASR
        & VIS & IR & @1 & @3 & ASR
        & VIS & IR & @1 & @3 & ASR
        & VIS & IR & @1 & @3 & ASR \\
        \midrule
        RandomPatch
        & 0.00 & 0.00 & 0.00 & 0.33 & 0.33
        & 0.00 & 0.00 & 0.00 & 0.33 & 0.00
        & 0.61 & 0.00 & 0.00 & 1.82 & 0.61
        & 0.00 & 0.00 & 0.00 & 1.82 & 0.91 \\
        TOUAP
        & 7.33 & 37.00 & 2.00 & 26.00 & 11.33
        & 43.67 & 12.00 & 9.00 & 19.67 & 35.33
        & 9.09 & 12.73 & 3.03 & 7.88 & 9.70
        & 28.64 & 19.55 & 10.00 & 19.55 & 27.73 \\
        UNIPatch
        & 1.00 & 3.67 & 1.00 & 8.67 & 4.00
        & 7.67 & 0.33 & 0.33 & 0.33 & 12.67
        & 1.82 & 4.85 & 0.61 & 5.45 & 4.24
        & 3.64 & 3.64 & 1.82 & 7.27 & 8.64 \\
        CDUPatch
        & 3.33 & 3.00 & 2.67 & 8.67 & 5.33
        & 0.00 & 0.33 & 0.00 & 6.00 & 0.67
        & 4.24 & 3.03 & 1.82 & 7.27 & 4.85
        & 7.27 & 4.55 & 4.09 & 19.55 & 7.27 \\
        LoSA
        & \textbf{45.33} & \textbf{45.00} & \textbf{44.67}
        & \textbf{54.00} & \textbf{45.00}
        & \textbf{73.00} & \textbf{71.00} & \textbf{71.00}
        & \textbf{79.67} & \textbf{76.33}
        & \textbf{30.91} & \textbf{29.09} & \textbf{29.09}
        & \textbf{39.39} & \textbf{30.91}
        & \textbf{91.36} & \textbf{90.00} & \textbf{90.00}
        & \textbf{94.09} & \textbf{91.82} \\
        \bottomrule
    \end{tabular}%
    }
\end{table*}

\begin{figure*}[t]
    \centering
    \includegraphics[width=\textwidth]{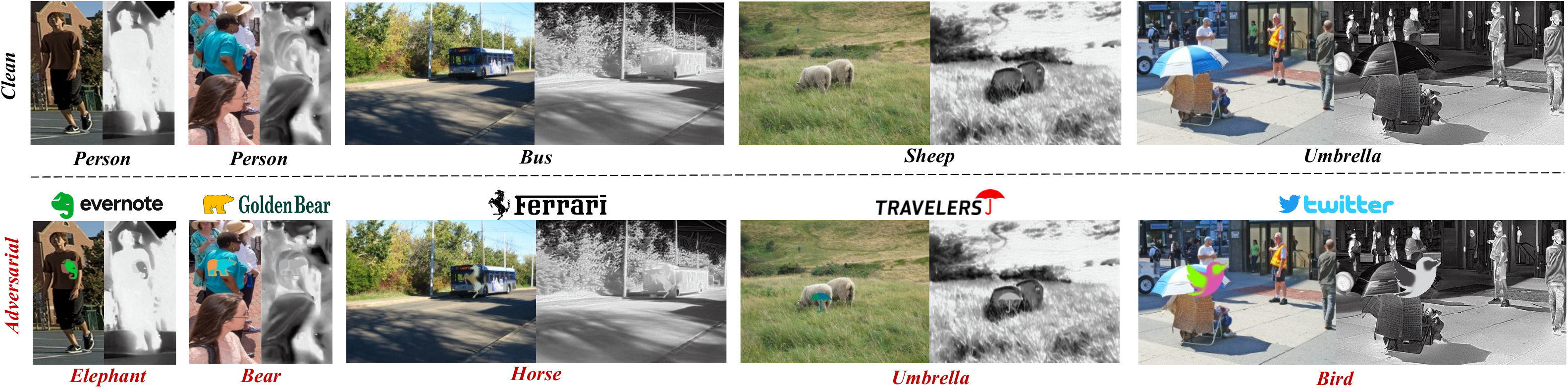}
    \caption{\textbf{Qualitative examples for zero-shot classification.} Clean (top) and adversarial (bottom) VIS--IR pairs are shown, with prescribed target categories below.}
    \label{fig:classification_examples}
\end{figure*}

\subsection{Experimental setup}
\label{sec:experimental_setup}

\noindent\textbf{Datasets.}
We evaluate on RGBDT500~\cite{zhu2025rgbdt500},
LasHeR~\cite{lasher}, COCO~\cite{lin2014coco},
and CrowdHuman~\cite{shao2018crowdhuman}.
RGBDT500 contributes 300 uncropped image pairs from nine video
sequences. We retain matching clean Top-1 predictions; both branches must
predict \texttt{person} for LasHeR and CrowdHuman, and the source
category for RGBDT500.
COCO uses 30 candidate classes; RGBDT500, LasHeR, and CrowdHuman
add \texttt{person}, yielding 31 candidates.
Five logo-to-target mappings are fixed before optimization.

\smallskip
\noindent\textbf{Models.}
Our main classification evaluation uses OpenAI CLIP ViT-L/14~\cite{radford2021clip}.
For cross-architecture evaluation, we optimize attacks independently on OpenCLIP ViT-B/16~\cite{cherti2022openclip}, MetaCLIP ViT-L/14~\cite{xu2023metaclip}, and EVA-CLIP EVA01-g/14-plus~\cite{sun2023evaclip}.

\begin{samepage}
\noindent Captioning and VQA transfer use six generative VLMs, comprising LLaVA-1.5/1.6~\cite{liu2023improved,liu2024llavanext}, OpenFlamingo~\cite{awadalla2023openflamingo}, BLIP-2 ViT-L and Flan-T5-XL~\cite{li2023blip2}, and InstructBLIP~\cite{dai2023instructblip}.
These tasks reuse classification adversarial images.
All models use frozen VIS and IR checkpoints.
\par
\end{samepage}

\smallskip
\noindent\textbf{Baselines.}
We compare LoSA with RandomPatch, classification adaptations of the VIS--IR patch attacks TOUAP~\cite{hu2023touap}, UNIPatch~\cite{wei2023unipatch}, and CDUPatch~\cite{long2025cdupatch}, using the same eligible pairs.

\smallskip
\noindent\textbf{Evaluation metrics.}
For classification, attack success rate (ASR) measures the percentage of eligible pairs whose adversarial Top-1 label differs from its clean prediction.
Targeted attack success rate (TASR@1) measures target Top-1 matches after excluding pairs whose clean Top-1 already equals the target.
We report each metric for VIS, IR, and their conjunction on the same pair (Both).
All pairs eligible for each metric enter its denominator.
For captioning and VQA, we use LLM-as-Judge~\cite{zheng2023judge} with GPT-6.1Sol to assess generated captions and answers.

\smallskip
\noindent\textbf{Implementation details.}
LoSA uses 20 particles for at most 200 iterations, evaluating one untransformed view per branch.

\begin{table*}[!t]
    \centering
    \caption{\textbf{Zero-shot attacks across CLIP architectures.} VIS/IR/Both report TASR@1; @3 and ASR refer to Both. Best values are bold.}
    \label{tab:cross_architecture}
    \small
    \setlength{\tabcolsep}{2pt}
    \begin{tabular*}{\textwidth}{@{\extracolsep{\fill}}l*{15}{c}@{}}
        \toprule
        & \multicolumn{5}{c}{OpenCLIP ViT-B/16} & \multicolumn{5}{c}{MetaCLIP ViT-L/14} & \multicolumn{5}{c}{EVA-CLIP EVA01-g/14-plus} \\
        \cmidrule(lr){2-6} \cmidrule(lr){7-11} \cmidrule(lr){12-16}
        Method & VIS & IR & Both & @3 & ASR & VIS & IR & Both & @3 & ASR & VIS & IR & Both & @3 & ASR \\
        \midrule
        RandomPatch & 0.00 & 0.32 & 0.00 & 2.10 & 0.96 & 1.28 & 0.30 & 0.00 & 1.15 & 1.92 & 0.00 & 0.56 & 0.00 & 2.82 & 0.83 \\
        TOUAP & 12.00 & 24.84 & 5.20 & 24.00 & 17.75 & 12.99 & 23.47 & 3.72 & 16.93 & 13.55 & 17.43 & 33.59 & 10.39 & 27.37 & 19.18 \\
        UNIPatch & 8.53 & 17.51 & 6.97 & 24.76 & 9.76 & 5.98 & 6.80 & 3.16 & 10.99 & 14.83 & 9.17 & 14.16 & 8.60 & 21.02 & 10.93 \\
        CDUPatch & 3.29 & 2.79 & 2.15 & 8.76 & 8.45 & 2.52 & 4.28 & 2.27 & 11.07 & 5.98 & 2.10 & 1.71 & 0.30 & 10.15 & 3.81 \\
        LoSA & \textbf{65.40} & \textbf{65.07} & \textbf{65.07} & \textbf{71.34} & \textbf{67.32} & \textbf{46.38} & \textbf{46.68} & \textbf{45.78} & \textbf{56.47} & \textbf{48.60} & \textbf{61.84} & \textbf{59.53} & \textbf{58.69} & \textbf{70.53} & \textbf{60.36} \\
        \bottomrule
    \end{tabular*}
\end{table*}

\subsection{Zero-shot classification effectiveness}
\label{sec:effectiveness}

We evaluate whether LoSA can induce the same prescribed target
prediction in both VIS and IR while preserving the supplied logo
outline.
As shown in \cref{tab:classification_effectiveness}, LoSA achieves
the highest simultaneous Top-1 targeted success among the evaluated
methods across all four datasets.
It also achieves the highest simultaneous Top-3 targeted success
and Both-ASR across these datasets, supporting its effectiveness
in jointly misleading the two modality branches under the stated
method-specific settings.
\Cref{fig:classification_examples} presents representative clean
and adversarial VIS--IR image pairs.

\subsection{Visual naturalness evaluation}
\label{sec:stealthiness}

\begin{figure}[!htbp]
    \centering
    \includegraphics[width=0.95\linewidth]{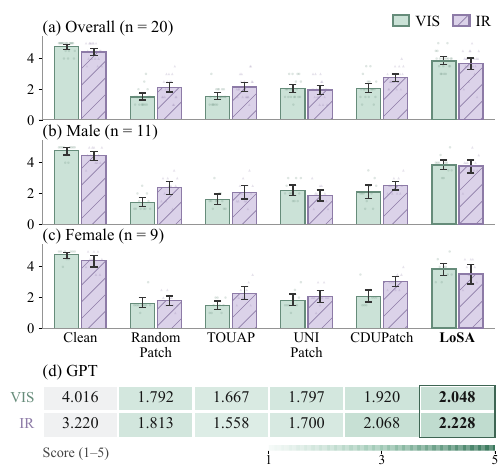}
    \caption{\textbf{Visual naturalness evaluation.}
    (a--c) Human ratings.
    (d) GPT-based naturalness scores.}
    \label{fig:human_naturalness}
\end{figure}

\begin{figure*}[!t]
    \centering
    \includegraphics[width=0.93\textwidth]{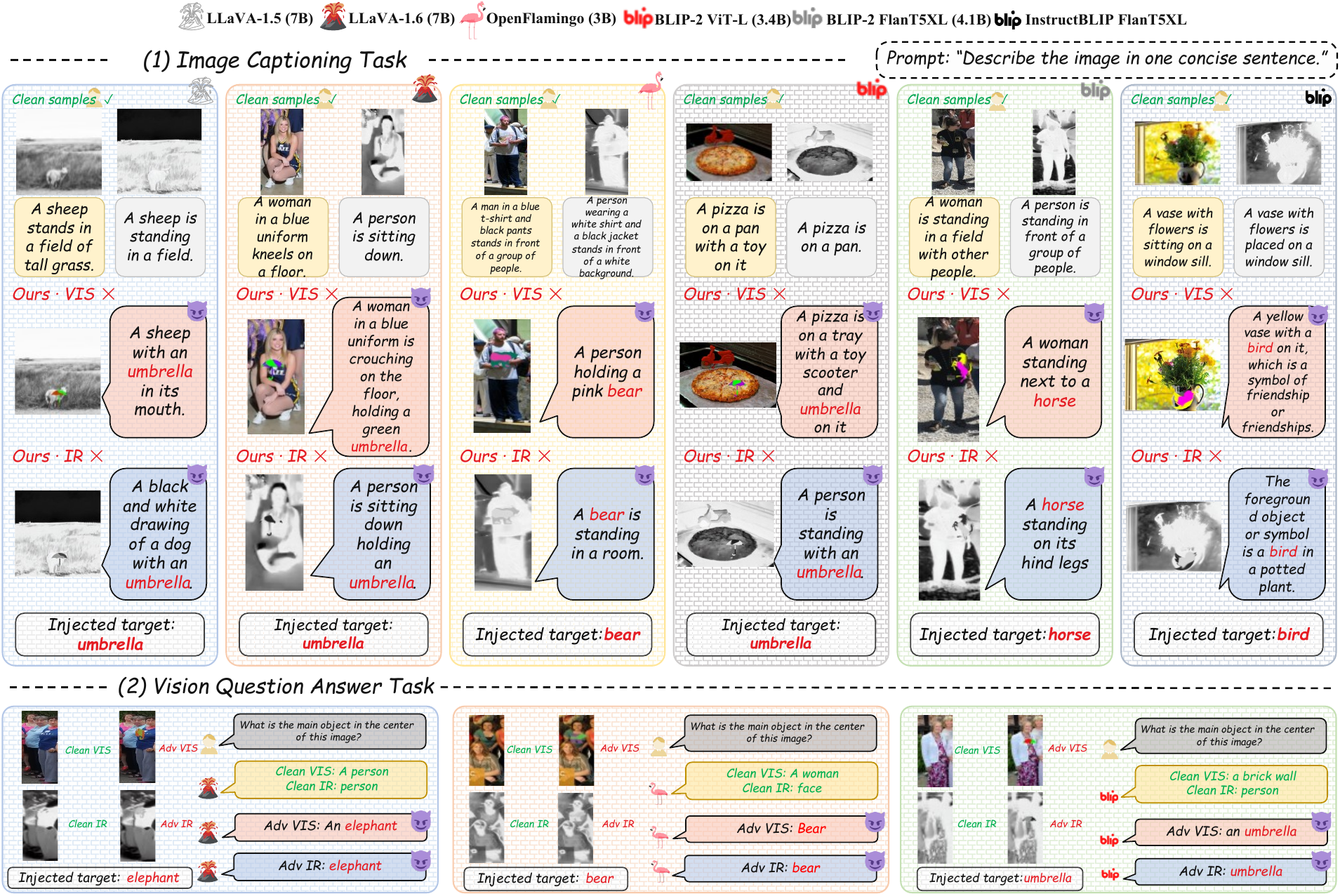}
    \caption{\textbf{Cross-task examples.} Classification-optimized images are reused; red words mark targets. Judging separates entities from printed motifs.}
    \label{fig:caption_vqa_examples}
    \par\smallskip
    \captionof{table}{\textbf{Captioning under transferred attacks.} Scores (0--100); parentheses: drops from Clean; bold: largest per model/modality.}
    \label{tab:transfer_caption_quality}
    \fontsize{8}{9.4}\selectfont
    \setlength{\tabcolsep}{1pt}
    \resizebox{\textwidth}{!}{%
    \begin{tabular}{@{}l*{12}{c}@{}}
        \toprule
        Model & \multicolumn{2}{c}{Clean} & \multicolumn{2}{c}{RandomPatch} & \multicolumn{2}{c}{TOUAP} & \multicolumn{2}{c}{UNIPatch} & \multicolumn{2}{c}{CDUPatch} & \multicolumn{2}{c}{LoSA} \\
        \cmidrule(lr){2-3} \cmidrule(lr){4-5} \cmidrule(lr){6-7} \cmidrule(lr){8-9} \cmidrule(lr){10-11} \cmidrule(lr){12-13}
         & VIS & IR & VIS & IR & VIS & IR & VIS & IR & VIS & IR & VIS & IR \\
        \midrule
        LLaVA-1.5 & 76.89 & 67.58 & 74.75 (2.14) & 65.56 (2.03) & 71.94 (4.95) & 60.08 (7.51) & 75.20 (1.69) & 61.54 (6.04) & 72.62 (4.27) & 63.15 (4.43) & \textbf{62.70 (14.19)} & \textbf{51.54 (16.04)} \\
        LLaVA-1.6 & 84.73 & 74.37 & 77.45 (7.28) & 72.89 (1.48) & 75.50 (9.23) & 68.58 (5.79) & 78.68 (6.04) & 70.10 (4.28) & 77.77 (6.96) & 72.16 (2.22) & \textbf{73.44 (11.29)} & \textbf{60.71 (13.66)} \\
        OpenFlamingo & 73.71 & 57.94 & 65.84 (7.88) & 55.68 (2.25) & 65.14 (8.57) & 55.26 (2.67) & 69.99 (3.72) & 51.85 (6.09) & 66.23 (7.48) & 51.81 (6.13) & \textbf{58.64 (15.07)} & \textbf{44.30 (13.64)} \\
        BLIP-2 ViT-L & 82.56 & 74.83 & 78.00 (4.56) & 72.68 (2.16) & 78.36 (4.20) & 71.14 (3.69) & 80.25 (2.31) & 72.31 (2.52) & 77.84 (4.73) & 71.78 (3.05) & \textbf{77.58 (4.99)} & \textbf{65.32 (9.51)} \\
        BLIP-2 T5-XL & 84.72 & 74.76 & 77.93 (6.79) & 69.44 (5.31) & 76.99 (7.72) & 67.52 (7.23) & 80.31 (4.41) & 66.22 (8.53) & 76.91 (7.81) & 67.52 (7.23) & \textbf{76.81 (7.91)} & \textbf{61.90 (12.85)} \\
        InstructBLIP & 80.31 & 75.19 & 68.65 (11.66) & 63.33 (11.85) & \textbf{65.23 (15.08)} & 58.51 (16.68) & 72.49 (7.82) & 60.83 (14.36) & 68.33 (11.98) & 61.72 (13.47) & 65.73 (14.58) & \textbf{52.08 (23.10)} \\
        \bottomrule
    \end{tabular}%
    }
    \par\smallskip
    \captionof{table}{\textbf{General VQA under transferred attacks.} Correctness (\%); parentheses: drops from Clean; bold: largest per model/modality.}
    \label{tab:transfer_vqa_quality}
    \fontsize{8}{9.4}\selectfont
    \setlength{\tabcolsep}{1pt}
    \resizebox{\textwidth}{!}{%
    \begin{tabular}{@{}l*{12}{c}@{}}
        \toprule
        Model & \multicolumn{2}{c}{Clean} & \multicolumn{2}{c}{RandomPatch} & \multicolumn{2}{c}{TOUAP} & \multicolumn{2}{c}{UNIPatch} & \multicolumn{2}{c}{CDUPatch} & \multicolumn{2}{c}{LoSA} \\
        \cmidrule(lr){2-3} \cmidrule(lr){4-5} \cmidrule(lr){6-7} \cmidrule(lr){8-9} \cmidrule(lr){10-11} \cmidrule(lr){12-13}
         & VIS & IR & VIS & IR & VIS & IR & VIS & IR & VIS & IR & VIS & IR \\
        \midrule
        LLaVA-1.5 & 63.17 & 44.59 & 61.51 (1.65) & 43.32 (1.27) & 61.01 (2.16) & 43.51 (1.08) & 61.90 (1.27) & 42.18 (2.42) & 61.45 (1.72) & 43.13 (1.46) & \textbf{58.78 (4.39)} & \textbf{37.53 (7.06)} \\
        LLaVA-1.6 & 72.65 & 51.84 & 72.26 (0.38) & 50.19 (1.65) & 71.44 (1.21) & 48.41 (3.44) & 71.82 (0.83) & 49.30 (2.54) & 72.46 (0.19) & 48.73 (3.12) & \textbf{70.80 (1.84)} & \textbf{47.07 (4.77)} \\
        OpenFlamingo & 43.96 & 37.15 & 40.65 (3.31) & 35.94 (1.21) & 41.09 (2.86) & 34.35 (2.80) & 42.88 (1.08) & 34.54 (2.61) & 41.22 (2.74) & 35.75 (1.40) & \textbf{37.09 (6.87)} & \textbf{31.55 (5.60)} \\
        BLIP-2 ViT-L & 61.96 & 43.77 & 59.54 (2.42) & 43.83 (-0.06) & \textbf{58.40 (3.56)} & \textbf{41.28 (2.48)} & 60.75 (1.21) & 42.68 (1.08) & 58.97 (2.99) & 42.49 (1.27) & \textbf{58.40 (3.56)} & 41.98 (1.78) \\
        BLIP-2 T5-XL & 66.51 & 49.03 & 63.78 (2.74) & 47.57 (1.47) & 61.91 (4.60) & 47.30 (1.73) & 63.84 (2.67) & 46.43 (2.60) & 62.71 (3.80) & 46.30 (2.74) & \textbf{59.37 (7.14)} & \textbf{43.76 (5.27)} \\
        InstructBLIP & 65.98 & 48.63 & 64.91 (1.07) & 48.30 (0.33) & 64.58 (1.40) & 46.43 (2.20) & 65.18 (0.80) & 47.10 (1.53) & 64.98 (1.00) & 47.36 (1.27) & \textbf{62.04 (3.94)} & \textbf{44.76 (3.87)} \\
        \bottomrule
    \end{tabular}%
    }
\end{figure*}

We assess the visual naturalness of optimized logos using human ratings and GPT-6.1Sol. As shown in \cref{fig:human_naturalness}(a--c), LoSA receives higher mean human ratings than the evaluated adversarial baselines in both modalities, overall and within each gender group. The GPT scores in \cref{fig:human_naturalness}(d) likewise rank LoSA highest among these baselines in both modalities. These results suggest improved relative visual plausibility on the evaluated samples.

\subsection{Cross-architecture evaluation}
\label{sec:cross_architecture}

We optimize attacks independently on OpenCLIP, MetaCLIP, and EVA-CLIP.
\Cref{tab:cross_architecture} shows higher simultaneous targeted success than the baselines on each architecture, using samples selected from the previously described datasets.

\section{Analysis}
\label{sec:analysis}

\subsection{Cross-task transferability}
\label{sec:cross_task}

We reuse classification adversarial images for captioning and VQA without further optimization.
LLaVA-1.5/1.6, OpenFlamingo, and BLIP-2 ViT-L use OpenAI CLIP ViT-L/14 image encoders; BLIP-2 Flan-T5-XL and InstructBLIP use EVA-CLIP ViT-G/14.

\Cref{tab:transfer_caption_quality,tab:transfer_vqa_quality} show reduced caption quality and general VQA correctness, with LoSA producing the largest drops in most settings.
\Cref{fig:caption_vqa_examples} presents selected outputs.

\begin{figure}[!t]
    \centering
    \includegraphics[width=\columnwidth]{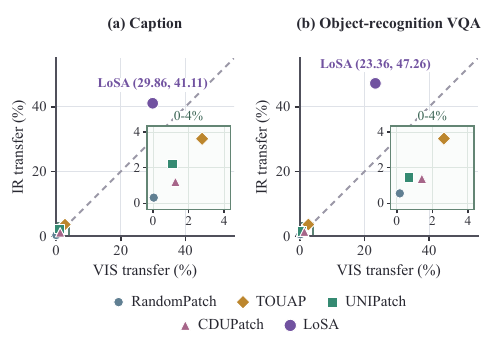}
    \caption{\textbf{Targeted semantic transfer.} Mean rates (\%) over six VLMs per modality; insets: baselines; diagonals: equal rates.}
    \label{fig:targeted_semantic_transfer}
\end{figure}

We further assess whether transferred attacks introduce the prescribed target as a scene entity.
In \cref{fig:targeted_semantic_transfer}, LoSA achieves the highest mean target-introduction rates: 29.86\%/41.11\% for captioning and 23.36\%/47.26\% for object-recognition VQA (VIS/IR). These results suggest that the learned target bias transfers beyond classification to captions and object-recognition answers, influencing semantic outputs without task-specific optimization.

\subsection{Ablation study}
\label{sec:ablation}

Ablations use OpenAI CLIP on samples selected from the previously described datasets, with a matched Full reference.

\begin{figure}[!t]
    \centering
    \begin{subfigure}[t]{0.48\columnwidth}
        \includegraphics[width=\linewidth]{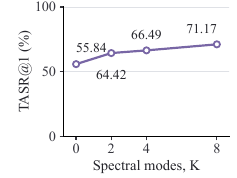}
        \caption{Spectral modes}
        \label{fig:ablation_spectral_modes}
    \end{subfigure}\hfill
    \begin{subfigure}[t]{0.48\columnwidth}
        \includegraphics[width=\linewidth]{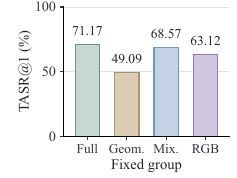}
        \caption{Optimization variables}
        \label{fig:ablation_geometry_color}
    \end{subfigure}
    \caption{\textbf{Field and optimization ablations.} Joint TASR@1 (\%). (a) Contour retained; $K=8$ is Full. (b) Geom./Mix./RGB fix geometry, gain/bias, or RGB endpoints.}
    \label{fig:ablation_summary}
\end{figure}

\noindent\textbf{Spectral modes and optimization variables.}
With the contour field retained, \cref{fig:ablation_spectral_modes} varies the number of spectral modes.
Joint TASR@1 rises from 55.84\% without modes to 71.17\% with eight, supporting richer interior variation within the fixed outline.
Fixing geometry, mixing controls, or RGB endpoints reduces the joint rate to 49.09\%, 68.57\%, or 63.12\%, respectively (\cref{fig:ablation_geometry_color}).
Fixing geometry has the largest effect among these controls.

\begin{table}[!htbp]
    \centering
    \caption{\textbf{Component ablations.} Pooled TASR@1 (\%) on OpenAI CLIP; Full is the matched reference.}
    \label{tab:ablation_structure}
    \small
    \setlength{\tabcolsep}{2pt}
    \begin{tabular*}{\columnwidth}{@{\extracolsep{\fill}}lccc@{}}
        \toprule
        Setting & VIS & IR & Both \\
        \midrule
        \textbf{Full} & \textbf{74.29} & \textbf{71.17} & \textbf{71.17} \\
        \midrule
        \multicolumn{4}{@{}l}{Field structure} \\
        w/o contour field & 68.57 & 65.97 & 65.97 \\
        w/o contour and modes & 58.96 & 50.65 & 50.39 \\
        Grid fill (3$\times$3 bilinear) & 69.61 & 65.97 & 65.97 \\
        \midrule
        \multicolumn{4}{@{}l}{Color initialization and texture balancing} \\
        w/o color initialization & 71.17 & 69.09 & 69.09 \\
        w/o texture balancing & 66.23 & 62.86 & 62.86 \\
        w/o initialization and balancing & 67.79 & 62.86 & 62.86 \\
        \midrule
        \multicolumn{4}{@{}l}{IR appearance} \\
        Uniform gray IR fill & 43.90 & 42.34 & 42.34 \\
        \bottomrule
    \end{tabular*}
\end{table}

\noindent\textbf{Field structure.}
In \cref{tab:ablation_structure}, removing the contour field lowers joint TASR@1 to 65.97\%, while removing both contour and spectral modes lowers it to 50.39\%.
These reductions support combining boundary-derived guidance with spectral variation.
To test a generic filling alternative, we replace the contour and spectral field with a $3\times3$ bilinear grid, retaining nine spatial coefficients and the 20-parameter optimization with the remaining settings unchanged.
This grid fill achieves 65.97\% joint TASR@1, 5.19 percentage points below Full, supporting the contour--spectral representation over this grid alternative.

\smallskip
\noindent\textbf{Color initialization and texture balancing.}
Removing color initialization yields 69.09\% joint TASR@1. Removing texture balancing, alone or together with initialization, yields 62.86\%, indicating greater sensitivity to texture balancing in this setting.

\smallskip
\noindent\textbf{IR appearance.}
Replacing the IR texture with uniform gray filling lowers joint TASR@1 to 42.34\%, supporting the use of variable IR appearance.

\subsection{Adversarial defense}
\label{sec:adversarial_defense}

We evaluate three digital-watermarking(DW)-inspired post-processing defenses~\cite{hayes2018visible} against fixed LoSA adversarial images.
DW1 uses the logo center, DW2 assumes no geometric priors, and DW3 uses its shape and size.

\begin{table}[!htbp]
    \centering
    \caption{\textbf{Attack success after defense.} Both requires success in both modalities on the same image pair.}
    \label{tab:adversarial_defense}
    \small
    \setlength{\tabcolsep}{2pt}
    \begin{tabular*}{\columnwidth}{@{\extracolsep{\fill}}l*{6}{c}@{}}
        \toprule
         & \multicolumn{3}{c}{ASR} & \multicolumn{3}{c}{TASR@1} \\
        \cmidrule(lr){2-4} \cmidrule(lr){5-7}
        Defense & VIS & IR & Both & VIS & IR & Both \\
        \midrule
        NoDefense & 67.63 & 78.68 & 65.79 & 66.05 & 64.47 & 64.47 \\
        DW1 & 65.26 & 75.00 & 60.53 & 63.42 & 59.47 & 56.84 \\
        DW2 & 60.00 & 63.95 & 46.58 & 56.58 & 43.16 & 38.95 \\
        DW3 & 54.74 & 49.21 & 31.84 & 51.84 & 26.84 & 21.84 \\
        \bottomrule
    \end{tabular*}
\end{table}

\Cref{tab:adversarial_defense} shows joint TASR@1 falling from 64.47\% to 56.84\%, 38.95\%, and 21.84\% under DW1--DW3, respectively.
Even under DW3, a subset of adversarial pairs continues to induce the prescribed target prediction in both modalities, indicating that the evaluated defenses provide only partial mitigation.

\FloatBarrier
\endgroup
\section{Conclusion}
\label{sec:conclusion}
We investigated whether familiar logos can serve as shape-preserving carriers for coordinated targeted semantic attacks on visible--infrared VLMs.
To this end, we introduced LoSA, which preserves a supplied logo's outline while jointly optimizing its interior appearance and placement through score-based black-box optimization.
Experiments on four datasets show higher simultaneous targeted success than the evaluated patch baselines, while cross-architecture, visual naturalness, and captioning/VQA analyses further characterize its behavior.
Future work will extend LoSA to physical-world settings, investigating its effectiveness and robustness under real-world imaging conditions.
Overall, the results identify logos as a distinct, shape-constrained attack surface for multimodal visual understanding.
\label{sec:main_text_end}

\begin{samepage}
\paragraph*{LLM Usage Statement.}
\label{sec:llm_usage}
ChatGPT was used for language editing and coding assistance. ChatGPT 6.1sol was used for automated visual naturalness assessment and for judging captioning and VQA outputs. The authors reviewed and verified the AI-assisted text, code, and evaluation results and take full responsibility for all content and findings.
\par
\end{samepage}
{
    \small
    \bibliographystyle{ieeenat_fullname}
    \bibliography{main}
}

\end{document}